\documentclass[11pt]{article}

\usepackage[final]{acl}

\usepackage{times}
\usepackage{latexsym}
\usepackage{forest}
\usepackage[T1]{fontenc}

\usepackage[utf8]{inputenc}

\usepackage{microtype}

\usepackage{inconsolata}
\usepackage{algorithm}
\usepackage{algpseudocode}
\usepackage{graphicx}
\usepackage[table,xcdraw]{xcolor}
\title{Can temporal article-level credibility signals improve domain-level credibility prediction?
}

 \author{Islam Eldifrawi \and Shengrui Wang \and Amine Trabelsi \\
     @usherbrooke.ca}

\begin{document}
\maketitle
\begin{abstract}
Web domain credibility evaluation is vital for combating misinformation. It is conducted by examining factors such as domain type, transparency, and overall reputation. However, assessing the credibility of newly emerging web domains remains challenging since they have no reputation yet. Expert fact-checkers evaluate the credibility of domains by analyzing the content of their articles, including the presence of misinformation, bias, or propaganda. Yet, the ease of large-scale content generation enabled by LLMs has accelerated the creation of new content, rendering manual assessment insufficient and underscoring the need for automated approaches to domain credibility evaluation. In this paper, we introduce our \textbf{D}omain \textbf{C}redibility \textbf{E}valuation \textbf{F}ramework (DCEF), a temporal framework for domain credibility evaluation grounded in expert ratings. DCEF enables us to investigate whether the credibility of web domains can be assessed from their published articles following the workflow of expert fact-checkers, without any prior knowledge of the source domains themselves.
\end{abstract}

\section{Introduction}

Assessing the credibility of the information source is vital in fact-checking \cite{eldifrawi-etal-2024-automated,guo-etal-2022-survey}. Due to recent advances in content generation, new domains are being created at an \textbf{unprecedented pace}, and most of them don't disclose their content generation methods. Manual evaluation of these domains credibility is infeasible. Automating the credibility assessment of these domains based on their content has become essential since they have no reputation yet. 

GNNs and pure graph-based approaches struggle to evaluate newly emerging domains with no historical data, or activity due to the “cold-start problem.” Since GNNs depend on connections between nodes, an isolated domain lacks the relational information needed for evaluation \cite{frej2024graph}.

Credibility (Believability) is assessed from content through a variety of heuristics known
as credibility signals, which help evaluate the overall trustworthiness of information sources \cite{leite2025weakly}. These signals include \textbf{the presence of bias, propaganda, misinformation, or misalignment between in article titles and their content.}

\begin{table}
\resizebox{\columnwidth}{!}{%
\begin{tabular}{|c|c|}
\hline
\rowcolor[HTML]{DAE8FC} 
Area       & \textbf{Open Gap}               \\ \hline
\rowcolor[HTML]{EFEFEF} 
Generalization  & Contextual adaptation \\ \hline
\rowcolor[HTML]{EFEFEF} 
Bias \& Fairness & Political/ ideological bias mitigation    \\ \hline
\rowcolor[HTML]{EFEFEF} 
Dynamics     & Temporal credibility evaluation        \\ \hline
\rowcolor[HTML]{EFEFEF} 
Insight  & Presence of model justification   \\ \hline
\end{tabular}%
}
\caption{Gaps in domain credibility evaluation models
\cite{srba2024survey}.}
\label{cred_gaps}
\end{table}

Credibility signals can be categorized as subjective or objective indicators\cite{srba2024survey}. Subjective indicators include aspects such as the design of a webpage, and domain popularity. While these features may influence user perception of credibility, they are not necessarily reliable measures of content quality. In contrast, objective indicators provide stronger evidence of quality and include the presence of bias, propaganda, misleading information, and the number of backlinks.

Recent deep learning models, like BERT-style transformers, when fine-tuned on misinformation corpora, often achieve high in-domain accuracy but tend to generalize poorly across different domains. For instance, a model trained on political news may perform poorly when applied to health misinformation \cite{srba2024survey}. In addition, they lack justifications for their classification.

LLMs have shown superior cross-domain performance compared to supervised baselines \cite{srba2024survey}. However, LLMs also have notable limitations. Their credibility judgments only moderately align with expert assessments (Spearman correlation nearly 0.50) \cite{10.1145/3717867.3717903}. In addition, they tend to assign higher credibility ratings to left-biased domains, as their pre-training data reflects a political bias leaning toward the left \cite{yang2025accuracy}. However, when they rate political bias, they are on par with the current Pre-trained Language Models (PLMs) SOTA models \cite{10.1145/3677389.3702601}.

Furthermore, the limitations extend beyond the models to the currently available datasets. Most existing datasets provide only domain URLs with credibility ratings, while only a limited number include links to domain articles. Moreover, a lot of these article links are no longer accessible.

Unlike professional credibility evaluators, most existing approaches do not take into account the temporal dynamics of domain credibility. \textbf{A domain may initially publish highly credible articles but experience a decline in quality over time, or conversely, it may improve its standards after a period of low credibility. Therefore, credibility ratings should place greater emphasis on recent articles}. To sum up, the credibility evaluation gaps are in Table \ref{cred_gaps}. In this paper, we investigate if domains credibility evaluation is improved through the incorporation of temporal article-level credibility signals. Our contribution is as follows:

\textbf{C1.} We construct a dataset that contains article content, addressing the issue of broken links in the current datasets. The dataset includes more than 300 domains and 25,141 articles spanning 14 fields.

\textbf{C2. }We introduce the DCEF framework. The input for DCEF is the time-stamped articles for a domain and the final output will be the credibility rating of the domain. \textbf{The framework will have no knowledge of the domain name it is evaluating, as the input will consist solely of a set of articles and their publication dates}. This addresses the 'Bias and Fairness' gap in Table \ref{cred_gaps}. DCEF is composed of \textbf{zero-shot} CoT LLM modules (to address the 'Generalization' gap in Table \ref{cred_gaps}). \textbf{The novelty here lies in the proposed architecture and its modular design}, providing justifications that address the 'Insight' Gap in Table \ref{cred_gaps}.

\textbf{C3. }We propose a novel methodology for incorporating temporal article-level features that are aggregated into the domain credibility classification process to address the 'Dynamics' gap in Table \ref{cred_gaps}. Furthermore, we evaluate their impact in Experiments \ref{FEM_tmp_exp}, \ref{ex2}, and \ref{cred_exp}.

\textbf{C4. }We study the influence of objective and subjective credibility indicators on domain credibility ratings in Experiment \ref{cred_exp}.

\section{Related Work}
Most recent work relies either on LLMs for textual credibility assessment or on graph-based approaches. Since GNNs and other graph-based methods cannot effectively assess the credibility of new domains with no historical footprint, our work focuses on LLM-based approaches.


\citet{10.1145/3717867.3717903}
audit
zero-shot approach by prompting nine LLMs. They evaluate their accuracy to assess the credibility of domains based on
their prior knowledge. The input is the domain name and the output is the credibility rating. They use the dataset from \cite{lin2023high}. They found out that larger models tend to refuse to rate lesser-known sources due to lack of information; smaller models are more likely to make errors; and LLMs show high agreement among themselves (Spearman’s = 0.79). However, they moderately align with human experts (0.50). 

Other works like \citet{mujahid-etal-2025-profiling} tried two approaches. Firstly, they tested zero-shot prompting multiple LLMs (GPT-3.5, Mistral-7B and LLama3-70B) to predict domain factuality and political bias by judging 5 recent articles per domain relying heavily on the LLMs prior knowledge. Then majority vote is applied on these 5 articles to produce the final domain factuality and bias classifications. Secondly, they try prompting gpt-3.5 per domain with 18 questions to collect some domain data guided by MBFC criteria, then use the generated LLM responses for downstream classification of bias with BERT and for factuality with SVM. Their second approach proved to be SOTA. 
However, their method doesn't address credibility classification, lacks temporal dynamics, and their data set of articles was not publicly released. 

Our approach evaluates domain credibility using the content and publication dates of articles within each domain without heavily relying on LLMs prior knowledge. 
\section{Methodology}

\begin{figure*}
  \centering
  \includegraphics[width=\textwidth]{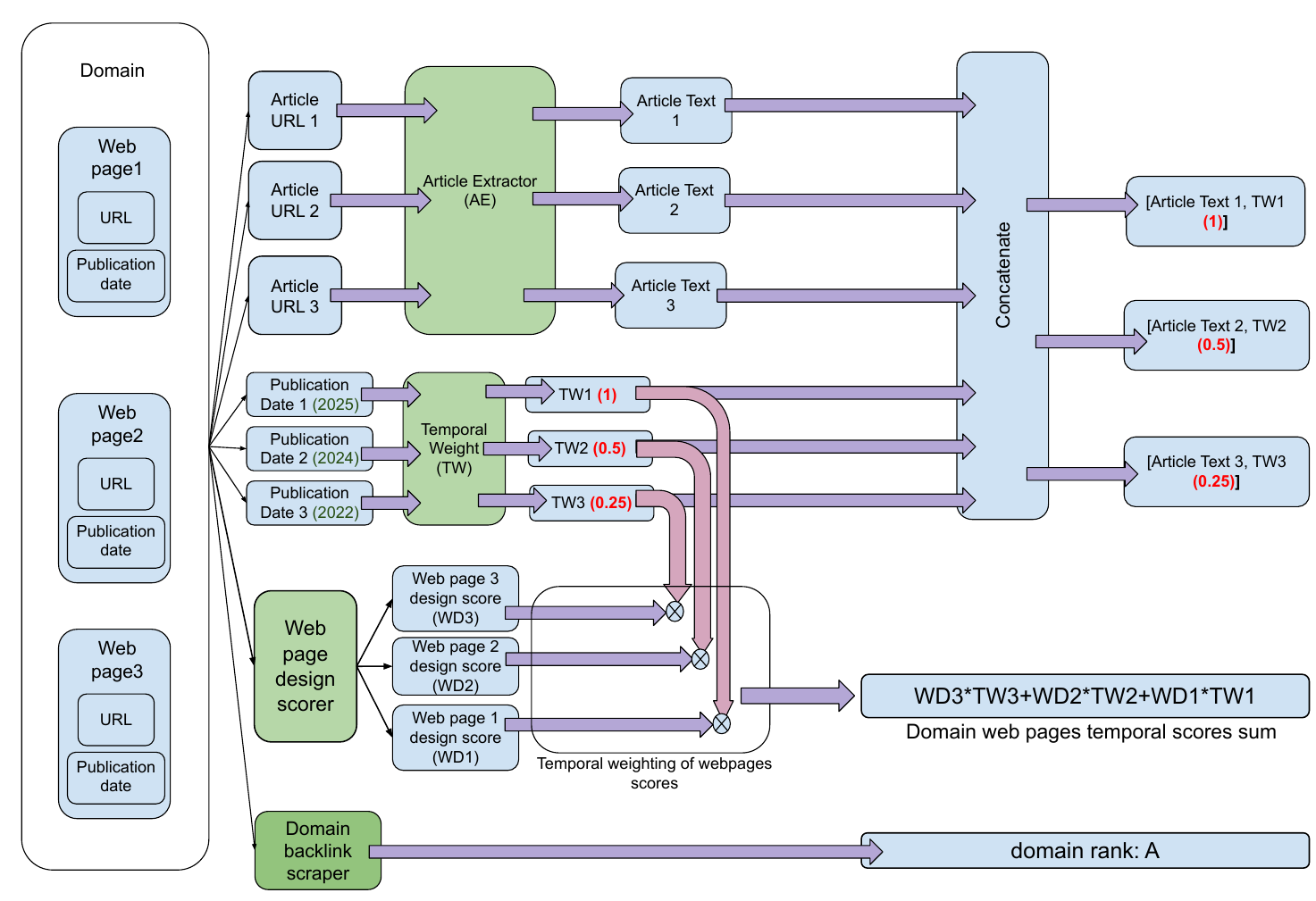}
  \caption{Stage 1 (The Preparation Stage): Prior to the evaluation stage, DCEF is inputted with the domain rank, articles URLs and their publication dates. DCEF scrapes the articles text. DECF also produces temporal weights using the publication date for each article and Equation \ref{eq:placeholder}. The result is a list of articles content and another list of their respective temporal weights generated from the corresponding article date of publication. Then each article is associated with its temporal weight. Examples of temporal weights based on publication dates  are shown in red.}
  \label{cred_ev1}
\end{figure*}

DCEF is composed of two stages: the preparation stage and the evaluation stage. In the preparation stage, domain articles and metadata are scraped to ensure that all information is available for the subsequent evaluation stage (see Figure \ref{cred_ev1}). The evaluation stage then processes this information to produce the final domain credibility ratings—low, medium, or high credibility— using a decision tree as it's known for its inherent interpretability as shown in Figure \ref{cred_ev2}. Each stage contains distinct modules, highlighted in green in both figures.

\subsection{The Preparation Stage}
\textbf{The input to the preparation stage consists of multiple articles URLs} from a given domain, \textbf{ with their publication dates}, however, the domain name is never given. 
The articles are then processed through a series of modules highlighted in green in Figure \ref{cred_ev1}, and described in detail below. 

\textbf{Article Extractor (AE):} We use Trafilatura \cite{barbaresi-2021-trafilatura} to extract the online important and relevant content from a given URL.

\textbf{Temporal Weight (TW): } Each article’s credibility features should be weighted by its publication date. The overall credibility of a domain is then computed as the temporally weighted sum of the credibility scores of its articles. For example, features of articles published in 2026 should carry more weight than those published in 2025. This weight is inversely proportional to the difference between an article’s publication year and that of the most recent article for the same domain. The temporal weight (TW) of an article is calculated using the following equation:
\begin{equation}
  (TW)_i = \frac{1}{y_r - y_i + 1}
  \label{eq:placeholder}
\end{equation}
where ${TW_i}$ is the temporal weight of the article credibility, ${y_r}$ is the year of the most recent article published in the domain, and ${y_i}$ is the year of publication of the target article. Note that adding 1 in the denominator prevents division by zero and maximizes the weights at 1. We considered using a neural network to learn the required weights or exponential decay with learnable parameters; however, this would introduce additional  complexity to DCEF and the weights might learn MBFC specific rating patterns decreasing its generalization capacity. Therefore, we opted to retain the current formula. Each article is associated with its temporal weight as shown in Figure \ref{cred_ev1}.

\textbf{Web page Design scorer (WD): }For every article, a range of features are evaluated including the text-to-HTML ratio, image-to-text ratio, hierarchy of headings, responsive design, average sentence length, font readability, and use of whitespace.

\textbf{Domain Ranker (DR): }
We extracted the number of backlinks and its rank for each domain from openlinkprofiler
 (an open source profiler)
 to provide an indication of the domain’s popularity. The ranks are "A+", "A", "B+", and "A-".

\subsection{The Evaluation Stage}
\begin{figure*}
  \centering
  \includegraphics[width=\textwidth]{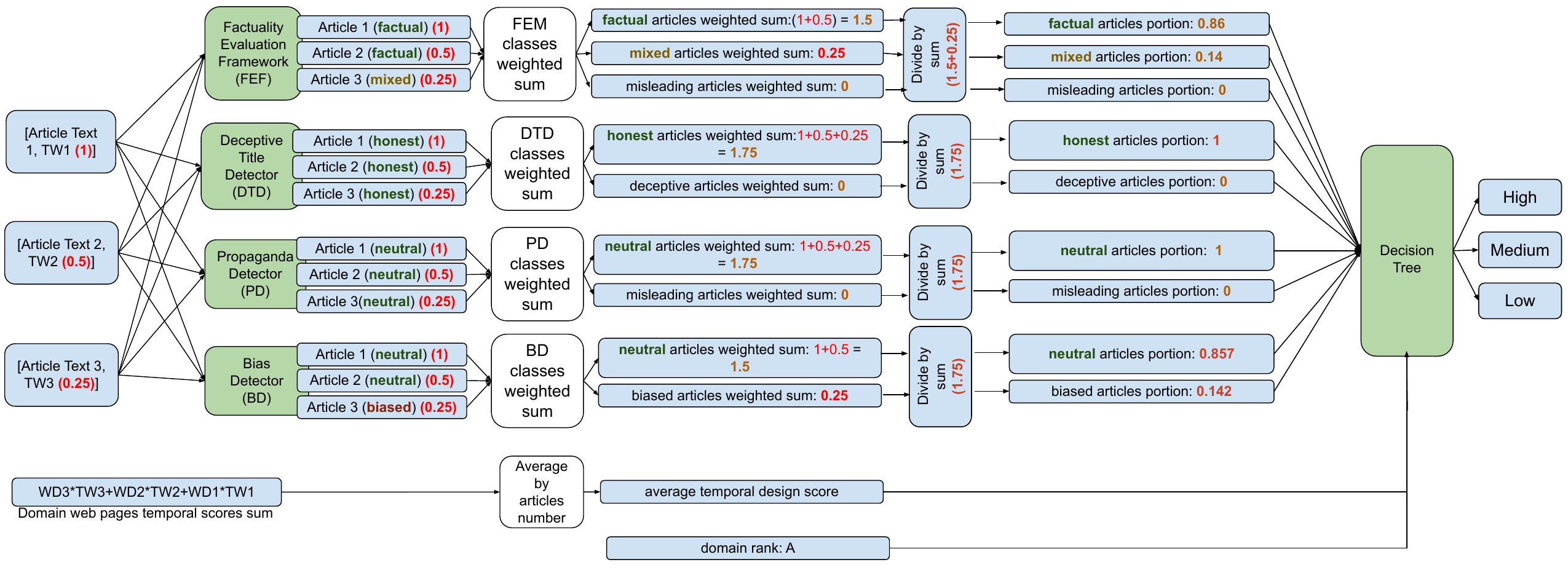}
  \caption{Stage 2 (The Evaluation Stage): DCEF evaluates the credibility of a domain by assessing each of its articles and aggregating the article-level evaluations into a final domain-level credibility rating. The numbers in red are examples of temporal weights of each article produced in the preparation stage. The article text is inputted in the four intermediate modules (FEM, DTD, PD and BD) and each article is categorized. The temporal weights of the articles are associated with the articles categories and aggregated using Algorithm \ref{algo} per module to produce temporal normalized features that will be inputted to the decision tree (final classifier).}
  \label{cred_ev2}
\end{figure*}

In this stage, the articles text obtained during the preparation stage is passed into four LLM modules (highlighted in green in Figure \ref{cred_ev2}) to generate the features required for domain credibility classification. These features are then aggregated, normalized and fed into a decision tree, which produces the final credibility rating. 

\textbf{Factuality Evaluation Module (FEM): } Fact-checking is the systematic process of verifying the factual accuracy of information, such as reports, public statements, or social media content. We scrape google search 
for retrieving evidence (information that will be our reference on judging the article veracity), 
then it is used to verify if the article is true, mixed (the article contains a combination of accurate and inaccurate/unverifiable information) or false based on the alignment with the evidence, and produce justification. The prompts are in the Appendix \ref{app_prompts} in Figures \ref{evidence_propmt} and \ref{fact_check_propmt}.

\textbf{Propaganda Detector (PD): } Propaganda detection is the process of identifying information, or communication deliberately designed to influence opinions, beliefs, or actions toward a specific, predetermined goal, often through manipulation, and selective or distorted information. For propaganda detection, the module produces one of two output classes—neutral or propaganda—along with a textual justification (see Appendix \ref{app_prompts} Figure \ref{propaganda_propmt} for the prompt detecting propaganda techniques). 

\textbf{Deceptive Title Detector (DTD):} The DTD assesses if the title accurately reflects the article’s content. The prompt shown in Appendix \ref{app_prompts} in Figure \ref{deceptive_propmt} is used to evaluate the alignment between an article’s title and its body text. The model is instructed to classify titles as honest or deceptive and to provide a brief justification.

\textbf{Bias Detector (BD): } Bias is considered an indicator of reduced credibility. To detect political bias, Llama-3.3-70B is prompted to classify the text as either neutral or biased, accompanied by a justification. The full prompts are provided in the Appendix \ref{app_prompts} Figure \ref{bias_propmt}. 

Various LLMs are tested as the backbone of the DCEF modules in Appendix \ref{LLM_sec_exp}, and Llama3.3-70B showed the best performance. After we get the classification from each module per article for the targeted domain, their temporal weights are grouped by their class to produce the temporal features. An example of the process done to produce the temporal features from the FEM is shown in Algorithm \ref{algo}. Same process is performed on the classes of BD, PD and DTD.
{\scriptsize
\begin{algorithm}

\caption{Computation of Temporal features from FEM per Domain}
\begin{algorithmic}[1]

\Require 
$factual\_weights$: list containing temporal weights of factual articles

\Require 
$mixed\_weights$: list containing temporal weights of mixed articles

\Require 
$misleading\_weights$: list containing temporal weights of misleading articles

\State $temp\_fact \gets \sum factual\_weights$
\State $temp\_mixed \gets \sum mixed\_weights$
\State $temp\_misleading \gets \sum mislead\_weights$

\State $normalizer \gets temp\_fact + temp\_mixed + temp\_mislead$

\State $temporal\_factual \gets \frac{temp\_fact}{normalizer}$

\State $temporal\_mixed \gets \frac{temp\_mixed}{normalizer}$

\State $temporal\_mislead \gets \frac{temp\_mislead}{normalizer}$
\State \Return
\Statex \hspace{\algorithmicindent}
$(temporal\_factual,$
$temporal\_mixed,$
\Statex \hspace{\algorithmicindent}
$temporal\_mislead)$\end{algorithmic}
\label{algo}
\end{algorithm}}

\textbf{The domain credibility classifier: }We selected a 5-levels decision tree with minimum 5 samples per leaf as the final classifier (See Appendix \ref{tree_choice} for more details about the tree selection). 

It is worth noting that LLMs are used in the classification of propaganda, bias, and factuality of \textbf{articles}, while decision trees will be used to classify the factuality, bias and credibility of \textbf{domains}. 

\section{Dataset Creation}
We constructed a balanced dataset to evaluate the credibility of 300 domains (100 low credibility, 100 medium and 100 high credibility domains) from MBFC based on their 25,141 articles. For each domain, we obtained credibility labels, bias categories (left, right, left-centered, right-centered, pro-science, least-biased), and factuality ratings (low, mixed, mostly factual, high, very high). There are 14 diverse topics in our dataset shown in Table \ref{topics}. We scraped the available English articles of these domains. Of these, the articles of 111 domains originate from five benchmark datasets-Politifact \cite{garg2020new}, MultiFC \cite{augenstein2019multifc}, PUBHEALTH \cite{kotonya2020explainable}, Credule \cite{chrysidis2024credible}, and FakeNewsCorpus \cite{pathak2019breaking}-while the articles of the remaining 189 were scraped from the web. Within the subset of the 189 domains, 60 domains (20 high credibility, 20 medium credibility, and 20 low credibility) contain articles that were not seen by the tested LLMs. Among these, 30 domains completely emerged entirely after the release of the tested LLMs, while the remaining 30 domains (despite being well known) consist exclusively of articles published after the models’ release dates. When given only the article texts, Llama 3.3 was unable to correctly infer the identities of the 60 source domains (see Appendix \ref{guess_dom}). Furthermore, each domain contains a different number of articles. We didn't try to change or increase the number of articles per domain in order to examine the impact of varying article counts on credibility assessment. On average, each article contains 567.17 words.

\begin{table}
\centering
\resizebox{5cm}{!}{%
\begin{tabular}{cr}
\rowcolor[HTML]{F2F2F2} 
{\color[HTML]{1F1F1F} \textbf{Politics \& Current Affairs}} & {\color[HTML]{1F1F1F} 14868} \\
\rowcolor[HTML]{FFFFFF} 
{\color[HTML]{1F1F1F} \textbf{Arts \& Entertainment}} & {\color[HTML]{1F1F1F} 1889} \\
\rowcolor[HTML]{F2F2F2} 
{\color[HTML]{1F1F1F} \textbf{Health \& Wellness}} & {\color[HTML]{1F1F1F} 1694} \\
\rowcolor[HTML]{FFFFFF} 
{\color[HTML]{1F1F1F} \textbf{Business \& Money}} & {\color[HTML]{1F1F1F} 1277} \\
\rowcolor[HTML]{F2F2F2} 
{\color[HTML]{1F1F1F} \textbf{Social Impact \& Voices}} & {\color[HTML]{1F1F1F} 811} \\
\rowcolor[HTML]{FFFFFF} 
{\color[HTML]{1F1F1F} \textbf{Sports}} & {\color[HTML]{1F1F1F} 548} \\
\rowcolor[HTML]{F2F2F2} 
{\color[HTML]{1F1F1F} \textbf{Style \& Beauty}} & {\color[HTML]{1F1F1F} 541} \\
\rowcolor[HTML]{FFFFFF} 
{\color[HTML]{1F1F1F} \textbf{Environment}} & {\color[HTML]{1F1F1F} 496} \\
\rowcolor[HTML]{F2F2F2} 
{\color[HTML]{1F1F1F} \textbf{Technology \& Science}} & {\color[HTML]{1F1F1F} 491} \\
\rowcolor[HTML]{FFFFFF} 
{\color[HTML]{1F1F1F} \textbf{Travel \& Home}} & {\color[HTML]{1F1F1F} 443} \\
\rowcolor[HTML]{F2F2F2} 
{\color[HTML]{1F1F1F} \textbf{Education and Religion}} & {\color[HTML]{1F1F1F} 415} \\
\rowcolor[HTML]{FFFFFF} 
{\color[HTML]{1F1F1F} \textbf{Relationships \& Family}} & {\color[HTML]{1F1F1F} 410} \\
\rowcolor[HTML]{F2F2F2} 
{\color[HTML]{1F1F1F} \textbf{Good and Weird News}} & {\color[HTML]{1F1F1F} 263} \\
\rowcolor[HTML]{FFFFFF} 
{\color[HTML]{1F1F1F} \textbf{Food and Drink}} & {\color[HTML]{1F1F1F} 251}
\end{tabular}%
}
\caption{Distribution of articles under each topic}
\label{topics}
\end{table}

The dataset contribution is in the unified labeling, de-duplication, and the augmentation of new content that has not been likely seen by Llama3.3.

\section{Experiments and Results}

To test the article-level signals produced by DCEF LLM modules, each module performance is evaluated against human annotations. The Pearson correlation between the majority vote of the three human annotators, and the BD, DTD, PD and FEM is 0.816, 0.96, 0.854 and 0.888 respectively. The experiment details are in Appendix \ref{hum_ann}. These significant correlations indicate that DCEF align reasonably well with human judgments.

Regarding testing the domain-level signals, we report 5-fold cross-validation results and the standard deviation (Std. Dev.) across the folds to ensure more stability. 
MBFC and many studies,\cite{vargas2023predicting, sanchez2024mapping, lin2023high, mujahid-etal-2025-profiling}, emphasize factuality and bias as the primary signals of credibility, while giving considerably less attention to other signals, such as propaganda and deceptive headlines. Therefore, We provide separate studies for the effect of temporal features on domain factuality and bias detection in Experiments \ref{FEM_tmp_exp} and \ref{ex2}. However, we evaluate the impact of \textbf{all} the temporal features like propaganda and deceptive titles on overall credibility assessment in Experiment \ref{cred_exp}. Although the domain factuality assessment and bias detection experiments each use separate decision trees, our aim isn't to propose new factuality or bias detection pipelines, but to assess the effectiveness of the temporal features on these two credibility signals. Lastly, we investigate if DCEF is capable of evaluating new domains likely unseen by llama3.3 in Experiment \ref{ex_n_dom}.

\subsection{Evaluating the temporal features impact on domain factuality ratings}
\label{FEM_tmp_exp}
This experiment evaluates the impact FEM features inputted to a decision tree in \textbf{two setups}, with and without the application of temporal weighting. As a baseline, we apply the method of \cite{mujahid-etal-2025-profiling} (SVM with TF-IDF) after modifying their prompts to follow the MBFC 5-level factuality labels on our dataset. It reported an accuracy of 0.849 and macro F1-score of 0.682. Most of their errors are in the 30 new domains likely because their method relies heavily on LLM’s prior knowledge. 

\textbf{Setup 1: Without temporal weighting: } Per article, FEM outputs three classes—factual, mixed, and misleading. For each domain, we obtain the \textbf{number of articles} classified as factual, mixed, and misleading, which serve as features. These features are used as input to a decision tree classifier to assign the domain factuality to one of five levels: very high, high, mostly factual, mixed, or low same as MBFC labels. The results are shown in Table \ref{FEM_no_temp}.

\begin{table}
\centering
\resizebox{6cm}{!}{%
\begin{tabular}{c|cc|c|}
\cline{2-4}
 & \multicolumn{1}{c|}{\cellcolor[HTML]{ECF4FF}precision} & \cellcolor[HTML]{ECF4FF}recall & \cellcolor[HTML]{ECF4FF}f1-score \\ \hline
\multicolumn{1}{|c|}{\cellcolor[HTML]{EFEFEF}macro avg} & \multicolumn{1}{c|}{0.70} & 0.71 & 0.71 \\ \hline
\multicolumn{1}{|c|}{\cellcolor[HTML]{EFEFEF}weighted avg} & \multicolumn{1}{c|}{0.90} & 0.90 & 0.90 \\ \hline
\multicolumn{1}{|c|}{\cellcolor[HTML]{EFEFEF}accuracy} & \multicolumn{2}{c|}{} & 0.90 \\ \hline
\end{tabular}%
}
\caption{5-fold cross-validation metrics of FEM features without temporal weighting. Std. Dev is 0.019.}
\label{FEM_no_temp}
\end{table}

\textbf{Setup 2: With temporal weighting: } Instead of the number of factual, or non-factual articles for each domain, we weight these numbers temporally and normalize them to produce FEM temporal normalized features as shown in Algorithm \ref{algo}, then we input them into a decision tree to predict the factuality of the domain and the results are in Table \ref{FEM_temp}. By comparing the results of the two setups, it is observed that temporal features performs better.

\begin{table}
\centering
\resizebox{6cm}{!}{%
\begin{tabular}{c|cc|c|}
\cline{2-4}
 & \multicolumn{1}{c|}{\cellcolor[HTML]{ECF4FF}precision} & \cellcolor[HTML]{ECF4FF}recall & \cellcolor[HTML]{ECF4FF}f1-score \\ \hline
\multicolumn{1}{|c|}{\cellcolor[HTML]{EFEFEF}macro avg} & \multicolumn{1}{c|}{0.78} & 0.82 & 0.80 \\ \hline
\multicolumn{1}{|c|}{\cellcolor[HTML]{EFEFEF}weighted avg} & \multicolumn{1}{c|}{0.93} & 0.95 & 0.94 \\ \hline
\multicolumn{1}{|c|}{\cellcolor[HTML]{EFEFEF}accuracy} & \multicolumn{2}{c|}{} & 0.95 \\ \hline
\end{tabular}%
}
\caption{5-fold cross-validation metrics of FEM features with temporal weights. Std. Dev is 0.014.}
\label{FEM_temp}
\end{table}
\textbf{This experiment shows that temporal features of domain articles improves the factuality classification of the domain.
} Before applying temporal weights, there were 9 misclassifications; after applying it, this number decreased to 4. Error examples of articles / domains misclassifications before and after temporal features are in Appendix \ref{FEM_ER}.


\subsection{Evaluating the temporal BD bias ratings}
\label{ex2}
Based on the MBFC bias ratings, we evaluated the performance of the BD module, with and without temporal weighting. However, for the purpose of credibility assessment, we are not concerned with distinguishing between left-leaning and right-leaning bias; instead, we focus on whether bias is present. To this end, the fine-grained MBFC \textbf{Domain} bias classes were aggregated: \textit{left-center} and \textit{right-center} into\textit{ less\_bias}; \textit{left}, \textit{extreme-left}, \textit{extreme-right} and \textit{right} were grouped into \textit{biased}; and \textit{least biased} and \textit{pro-science} into \textit{neutral}.

\textbf{Setup 1: Without temporal weighting: } For each domain, we obtain the number of articles classified as neutral and the number of articles classified as biased. These counts are used as input for a decision tree classifier to predict the domain’s credibility. The results are in Table \ref{BD_no_temp}.

\begin{table}
\centering
\resizebox{6cm}{!}{%
\begin{tabular}{c|cc|c|}
\cline{2-4}
 & \multicolumn{1}{c|}{\cellcolor[HTML]{ECF4FF}precision} & \cellcolor[HTML]{ECF4FF}recall & \cellcolor[HTML]{ECF4FF}f1-score \\ \hline
\multicolumn{1}{|c|}{\cellcolor[HTML]{EFEFEF}macro avg} & \multicolumn{1}{c|}{0.80} & 0.84 & 0.81 \\ \hline
\multicolumn{1}{|c|}{\cellcolor[HTML]{EFEFEF}weighted avg} & \multicolumn{1}{c|}{0.84} & 0.82 & 0.83 \\ \hline
\multicolumn{1}{|c|}{\cellcolor[HTML]{EFEFEF}accuracy} & \multicolumn{2}{c|}{} & 0.82 \\ \hline
\end{tabular}%
}
\caption{5-fold cross-validation metrics of BD features without temporal weighting. Std. Dev is 0.006.}
\label{BD_no_temp}
\end{table}


\textbf{Setup 2: Incorporating factuality: }We added the numbers of factual, mixed and misleading articles to a similar decision tree without temporal weighting. Surprisingly, the performance metrics increased as shown in Table \ref{BD_temp}.
\begin{table}
\centering
\resizebox{6cm}{!}{%
\begin{tabular}{c|cc|c|c|}
\cline{2-4}
 & \multicolumn{1}{c|}{\cellcolor[HTML]{ECF4FF}precision} & \cellcolor[HTML]{ECF4FF}recall & \cellcolor[HTML]{ECF4FF}f1-score \\ \hline
\multicolumn{1}{|c|}{\cellcolor[HTML]{EFEFEF}macro avg} & \multicolumn{1}{c|}{0.88} & 0.91 & 0.89 \\ \hline
\multicolumn{1}{|c|}{\cellcolor[HTML]{EFEFEF}weighted avg} & \multicolumn{1}{c|}{0.91} & 0.90 & 0.90 \\ \hline
\multicolumn{1}{|c|}{\cellcolor[HTML]{EFEFEF}accuracy} & \multicolumn{2}{c|}{} & 0.90 \\ \hline
\end{tabular}%
}
\caption{5-fold cross-validation metrics for BD features with factual features without temporal weighting. Std. Dev is 0.005.}
\label{BD_temp}
\end{table}
\textbf{Further analysis of the relationship between factuality and bias shows that bias detection can benefit from factuality, as shown in Table \ref{bias_imp} for studying the feature importance, where the second important feature in bias detection is the number of mixed articles.} Factuality plays a role in bias detection on the article level, for example, PolitiFact published Trump’s claim that unemployment rates fell below 2\%. The article described the claim as \textbf{"absolutely nonsense"}. Since Trump’s claim was false, the statement doesn't constitute bias. However, if the claim was true, using \textbf{"absolutely nonsense"} would represent loaded language and bias.

\begin{table}
\centering
\resizebox{6cm}{!}{%
\begin{tabular}{|c|c|}
\hline
\rowcolor[HTML]{96FFFB} 
\textbf{Feature} & \textbf{Importance} \\ \hline
\rowcolor[HTML]{F7F7F7} 
{\color[HTML]{1F1F1F} temporal\_neutral\_articles} & {\color[HTML]{1F1F1F} 0.140185} \\ \hline
\rowcolor[HTML]{FFFFFF} 
{\color[HTML]{1F1F1F} \textbf{temporal\_biased\_articles}} & {\color[HTML]{1F1F1F} \textbf{0.493885}} \\ \hline
\rowcolor[HTML]{F7F7F7} 
{\color[HTML]{1F1F1F} temporal\_factual} & {\color[HTML]{1F1F1F} 0.078757} \\ \hline
\rowcolor[HTML]{FFFFFF} 
{\color[HTML]{1F1F1F} \textbf{temporal\_mixed}} & {\color[HTML]{1F1F1F} \textbf{0.269210}} \\ \hline
\rowcolor[HTML]{EEEEEE} 
{\color[HTML]{1F1F1F} temporal\_misleading} & {\color[HTML]{1F1F1F} 0.017964} \\ \hline
\end{tabular}%
}
\caption{Feature importance table for bias classification}
\label{bias_imp}
\end{table}
\textbf{Setup 3: Incorporating normalized temporal features and factuality: }We decided to input the temporal FEM with the temporal BD features in the decision tree. The 5-fold cross-validation metrics increased as shown in Table \ref{BD_tempo}. This indicates that using temporal features of bias and factuality improves domain bias detection. Examples of the impact of temporal features on reducing misclassifications is in Appendix \ref{bd_temp_err}. Compared to the BERT-based bias detection method of \cite{mujahid-etal-2025-profiling}, which achieved 0.89 accuracy across five runs and struggled mainly with the 30 new domains, our approach using temporal features with factuality achieved a higher accuracy of 0.92.

\begin{table}
\centering
\resizebox{6cm}{!}{%
\begin{tabular}{c|cc|c|}
\cline{2-4}
 & \multicolumn{1}{c|}{\cellcolor[HTML]{ECF4FF}precision} & \cellcolor[HTML]{ECF4FF}recall & \cellcolor[HTML]{ECF4FF}f1-score \\ \hline
\multicolumn{1}{|c|}{\cellcolor[HTML]{EFEFEF}macro avg} & \multicolumn{1}{c|}{0.89} & 0.92 & 0.90 \\ \hline
\multicolumn{1}{|c|}{\cellcolor[HTML]{EFEFEF}weighted avg} & \multicolumn{1}{c|}{0.91} & 0.91 & 0.91 \\ \hline
\multicolumn{1}{|c|}{\cellcolor[HTML]{EFEFEF}accuracy} & \multicolumn{2}{c|}{} & 0.92 \\ \hline
\end{tabular}%
}
\caption{5-fold cross-validation metric evaluating the temporal BD and factual features. Std. Dev is 0.005}
\label{BD_tempo}
\end{table}
\subsection{Domain credibility evaluations with DCEF}
\label{cred_exp}
We examine the impact of both subjective features-such as the design of a domain’s web pages-and objective features-such as the domain’s factuality-on the MBFC domain credibility ratings by analyzing feature importance. All features are provided as input to a decision tree to estimate the contribution of each feature and its importance. Table \ref{feat_imp} shows that\textbf{ the most important features are objective.}

\begin{table}
\centering
\resizebox{\columnwidth}{!}{%
\begin{tabular}{|c|r|}
\hline
\rowcolor[HTML]{DAE8FC} 
\textbf{Feature} & \multicolumn{1}{c|}{\cellcolor[HTML]{DAE8FC}\textbf{Importance}} \\ \hline
\rowcolor[HTML]{F7F7F7} 
{\color[HTML]{1F1F1F} \textbf{temporal factual articles}} & {\color[HTML]{1F1F1F} \textbf{0.567908}} \\ \hline
\rowcolor[HTML]{FFFFFF} 
{\color[HTML]{1F1F1F} temporal mixed articles} & {\color[HTML]{1F1F1F} 0.005088} \\ \hline
\rowcolor[HTML]{F7F7F7} 
{\color[HTML]{1F1F1F} \textbf{temporal misleading articles}} & {\color[HTML]{1F1F1F} \textbf{0.051756}} \\ \hline
\rowcolor[HTML]{FFFFFF} 
{\color[HTML]{1F1F1F} temporal neutral articles} & {\color[HTML]{1F1F1F} 0.006967} \\ \hline
\rowcolor[HTML]{F7F7F7} 
{\color[HTML]{1F1F1F} \textbf{temporal biased articles}} & {\color[HTML]{1F1F1F} \textbf{0.043129}} \\ \hline
\rowcolor[HTML]{FFFFFF} 
{\color[HTML]{1F1F1F} temporal title honest articles} & {\color[HTML]{1F1F1F} 0.008346} \\ \hline
\rowcolor[HTML]{F7F7F7} 
{\color[HTML]{1F1F1F} \textbf{temporal title deceptive articles}} & {\color[HTML]{1F1F1F} \textbf{0.012944}} \\ \hline
\rowcolor[HTML]{FFFFFF} 
{\color[HTML]{1F1F1F} temporal propagandistic articles} & {\color[HTML]{1F1F1F} 0.005653} \\ \hline
\rowcolor[HTML]{F7F7F7} 
{\color[HTML]{1F1F1F} \textbf{temporal non-propagandistic articles}} & {\color[HTML]{1F1F1F} \textbf{0.294327}} \\ \hline
\rowcolor[HTML]{FFFFFF} 
{\color[HTML]{1F1F1F} temporal domain layout design score} & {\color[HTML]{1F1F1F} 0.000000} \\ \hline
\rowcolor[HTML]{EEEEEE} 
{\color[HTML]{1F1F1F} domain rank} & {\color[HTML]{1F1F1F} 0.003881
} \\ \hline
\end{tabular}%
}
\caption{Feature importance after being fitted on a decision tree for domain credibility classification.} 
\label{feat_imp}
\end{table}


The temporal features identified as most important in Table \ref{feat_imp} were evaluated by comparing two setups: using all features versus using only the top five. The 5-fold cross-validation results show identical performance in both cases, with an accuracy and a macro F1-score of 0.884. 
This indicates that the top five features are sufficient. When we used the non-temporal features, performance decreased. The 5-fold cross-validation accuracy and macro f1-score became 0.834 with standard deviation of 0.007. Due to limited prior work on predicting credibility ratings, we prompt GPT-5.5 (Figure \ref{gpt_cred}) as our baseline to rate the credibility of domains based on 5 articles (like \cite{mujahid-etal-2025-profiling}) or less (3 or 1) depending on the number of available articles in our dataset. Then we apply majority voting. This baseline average accuracy and F1-score across 5 runs is 0.589 with standard deviation of 0.024 which aligns with \citet{10.1145/3717867.3717903}.
 
\begin{figure}
  \centering
  \includegraphics[width=6cm]{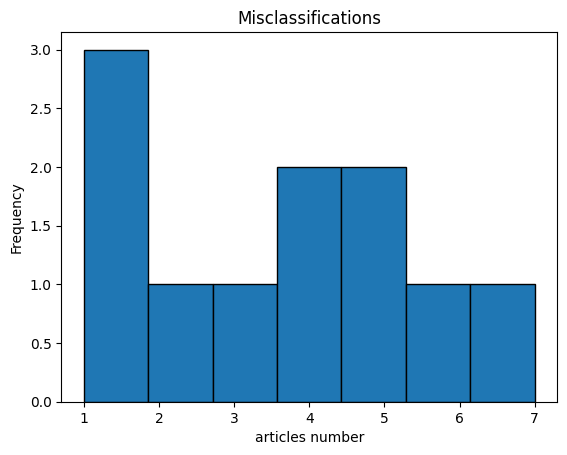}
  \caption{Relationship between the frequency of misclassifications and the number of articles per domain. }
  \label{mist_hist}
\end{figure}

Our error analysis in one of the 5 folds revealed two dominant patterns in 11 misclassifications. The first occurs when the underlying LLM produces hallucinations in evaluating factuality, bias, propaganda, or deceptive titles of articles in domains with a relatively small number of articles. In such cases, errors in just one recent article can significantly impact the domain-level classification. For example, consider the domain act.tv, which contains only five articles. In this case, a hallucination occurred in the factuality assessment of the article titled “Jon Stewart Says Blame for Trump's Win Goes Beyond Party Lines”, published in 2025. This article should have been labeled as mixed, but FEM incorrectly classified it as factual. Since it is the most recent article, it received a temporal weight of 0.71, while the other four articles—published in 2016—each received a weight of 0.071. Due to this imbalance, the misclassification of a temporal highly weighted article significantly influenced the prediction, shifting it from medium to high. The second pattern occurred in domains such as foxnews.com that lacked a sufficient representative sample of articles. According to MBFC, Fox News is rated as low credibility. Upon revisiting the data, we found that our dataset contained only a single article from Fox News, which was neutral, factual, and unbiased. Consequently, the tree classifier predicted the domain to be highly credible. This error does not reflect a weakness in the tree classifier; rather, it highlights the importance of having a sufficient number of articles per domain. In\textbf{ both error patterns, the number of input articles is low}. Figure \ref{mist_hist} illustrates the relationship between the frequency of misclassifications and the number of articles per domain. No misclassifications occurred with domains containing 8 articles or more. Based on this observation, we recommend more than 8 most recent articles to decrease the probability of DCEF misclassifications. Error analysis with examples of articles and domain misclassifications are in Appendix \ref{kokobt}.

\subsection{Testing DCEF on new content likely unseen by its underlying LLMs}
\label{ex_n_dom}
\begin{figure}
  \centering
  \includegraphics[width=7.5cm]{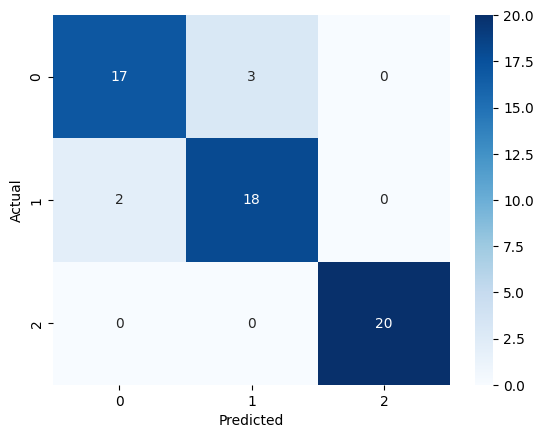}
  \caption{Confusion matrix on new content for with classes; low (0), medium (1), and high credibility (2)}
  \label{conf_mat}
\end{figure}

In this experiment, we collect all 60 domains (20 low, 20 medium and 20 high credible domains) containing articles published after the release of Llama 3.3-70B and assign them to the test set, while all remaining domains are used for training. As a baseline, we prompted GPT-5.5 to classify the credibility of these sixty domains based on its knowledge, it provided an accuracy and F1-score of 0.557 with standard deviation of 0.011. For DCEF, we only used the top five temporal features in Table \ref{feat_imp}. Notably, DCEF achieved an accuracy and F1-score of 0.92 on content that had never been seen by the underlying LLM. The strong performance in Table \ref{new_dom} and in the confusion matrix in Figure \ref{conf_mat}, demonstrate that automated credibility assessment can be effectively performed on newly emerging domains. The resulting decision tree is presented in Figure \ref{cred_tree}. 
The same error patterns observed in Experiment \ref{cred_exp} are also present here, indicating that the performance remains consistent regardless of whether the underlying LLM in DCEF has previously encountered the article. This suggests that the framework is not significantly influenced by prior exposure to the evaluated content. The 5 misclassifications in Figure \ref{conf_mat} are analyzed in Appendix \ref{errrr}.

\begin{table}
\centering
\resizebox{6cm}{!}{%
\begin{tabular}{c|cc|c|}
\cline{2-4}
 & \multicolumn{1}{c|}{\cellcolor[HTML]{ECF4FF}precision} & \cellcolor[HTML]{ECF4FF}recall & \cellcolor[HTML]{ECF4FF}f1-score \\ \hline
\multicolumn{1}{|c|}{\cellcolor[HTML]{EFEFEF}macro avg} & \multicolumn{1}{c|}{0.92} & 0.92 & 0.92 \\ \hline
\multicolumn{1}{|c|}{\cellcolor[HTML]{EFEFEF}weighted avg} & \multicolumn{1}{c|}{0.92} & 0.92 & 0.92 \\ \hline
\multicolumn{1}{|c|}{\cellcolor[HTML]{EFEFEF}accuracy} & \multicolumn{2}{c|}{} & 0.92 \\ \hline
\end{tabular}%
}
\caption{classification report for temporal weighting on the content likely unseen by Llama3.3}
\label{new_dom}
\end{table}

\section{Conclusion}
We introduce DCEF, a temporal framework for domain credibility evaluation. Our findings show that domain credibility expert ratings rely on objective temporal credibility indicators, and that bias detection improves when misinformation detection is inputted. We also provide a dataset of 25,141 articles enabling further study of the credibility of domains based on its articles.

\section*{Limitations}

\begin{enumerate}
  \item In this study, we used all available English articles from established datasets. Because many article URLs were inaccessible, the number of retrieved articles varied substantially across domains, with some domains represented by only a few articles and others by hundreds. Our primary goal was to examine whether temporal article-level signals can support credibility assessment for emerging domains under realistic data constraints. That said, we acknowledge that representative sampling is important for capturing the full distribution of a domain’s content. Future work will explore automated sampling methods, potentially guided by topic modeling, as well as uncertainty-aware aggregation to estimate whether the numer of available articles are sufficient for reliable domain-level inference. Addressing representative sampling remains an important direction, but it is beyond the scope of the present study.
  
  \item We use MBFC credibility ratings as a practical source-level proxy because they provide broad coverage, a transparent public methodology, and labels that are widely used in computational studies of media credibility and bias. Prior work has found substantial correspondence between MBFC and other news-quality rating systems, which suggests that the ratings capture a meaningful and reusable signal for comparative analysis. For instance, \citet{lin2023high}, discovered that there is a significant pearson correlation (0.81) between MBFC and Newsguard domain ratings. However, MBFC ratings reflect a specific editorial framework and may not fully capture topic-specific variation, or all forms of credibility relevant to every study. That is why we included propaganda and title deception as additional signals although they were less emphasized by MBFC. Furthermore, we don't rely solely on MBFC ratings, as we do human validation experiments on credibility, factuality and bias in Appendix \ref{hum_ann}. In addition, we avoided putting any learnable parameters in Equation \ref{eq:placeholder} that might be influenced by MBFC specific editorial framework.
  \item The use of Google search as evidence retrieval makes the factuality module hard to reproduce. However, the up-to-date evidence retrieval in realistic fact-checking settings, is inherently external and must be retrieved dynamically. We preferred this open-source alternative over commercial options such as Serper.
  \item This paper didn't put into consideration multi-lingual domain credibility evaluation and focused on English articles only. In future work, multi-lingual evaluations will be performed.
\end{enumerate}


\bibliography{custom}

\appendix
\section{Appendix}
All experiments were conducted on a Quadro RTX 8000 GPU, with the temperature of all LLMs set to zero.

\subsection{Decision Tree as a classifier Design choices and concerns about overfitting}
\label{tree_choice}
We opted for an interpretable classifier, as our primary objective is to understand and explain the classifier's final decisions. To this end, we considered three candidate approaches: decision trees, linear models, and logistic regression classifiers. Decision trees generally surpass linear or logistic regression when the underlying data has non-linear relationships, complex feature interactions, or requires minimal data preprocessing. While regression models assume a straight-line (or linear log-odds) relationship, decision trees split data into smaller, manageable subspaces to identify non-monotonic patterns \footnote{\url{https://gustavwillig.medium.com/decision-tree-vs-logistic-regression-1a40c58307d0}}.

Key scenarios where decision trees are superior:\footnote{\url{https://www.researchgate.net/post/Should_I_use_a_decision_tree_or_logistic_regression_for_classification}} 

\begin{itemize}
  \item Non-Linear Relationships. When the relationship between predictor variables and the outcome can't be described by a straight line, decision trees can effectively capture these patterns, whereas linear models will fail.
  \item Complex Feature Interactions: Decision trees naturally detect interaction effects (e.g., if feature X and feature Y work together to produce an outcome), whereas logistic regression requires explicit, manual creation of interaction terms.
  \item Irregular Data Distributions \& Outliers: Decision trees are robust to outliers, as splits are based on sample proportions within leaf nodes rather than absolute numerical values that affect line fitting in regression.Mixed Data Types and Preprocessing: Trees handle categorical and numerical features effortlessly without needing normalization, standardization, or dummy encoding (one-hot encoding).
  \item High-Dimensional Sparse Data: In datasets where many features are zero or missing, decision trees can work effectively by partitioning data based on feature presence.
\end{itemize}

A perfectly balanced binary tree of depth 5 has 32 leaf nodes. With 300 samples, each leaf would have an average of about 9.3 samples. This is enough to provide statistical significance for each leaf's prediction, especially if a Decision Tree Classifier with a min\_samples\_leaf setting of 5 or more is used to ensure robust decisions and decrease the probability of overfitting in trees\footnote{\url{https://medium.com/data-science/how-to-tune-a-decision-tree-f03721801680}}.

\subsection{Error Analysis}
\label{err_ana}
In this section we do deeper error analysis on the misclassified domains regarding their credibility evaluation. 
\subsubsection{ Error analysis before and after using temporal features with FEM}
\label{FEM_ER}
Before applying temporal weighting, there were 9 misclassifications; after incorporating it, this number decreased to 4. The analysis of the corrected cases shows that temporal weighting aligns the classifications more with MBFC ratings specially with domains that have articles with different publication years. For example, consider the domain cbssports.com. In our sample, there are 16 factual articles, 1 mixed article with title (What to watch: Baby Los Angeles Lakers play host to surging Golden State Warriors), and 1 misleading article with title (The Ravens have an interesting idea to change the NFL's kickoff rule). This corresponds to approximately 0.88 (16/18) factual content, and about 0.055 (1/18) each for mixed and misleading content. Based on these non-temporal features, the domain was classified as 'mostly factual'. After applying temporal weights, the proportions shifted: the mixed and misleading articles decreased to 0.05 each, based on Equation \ref{eq:placeholder} and Algorithm \ref{algo} as they were published in 2016, while the proportion of factual articles increased to 0.9, since they were published between 2017 to 2020. Thus, the classification changed to 'high' similar to MBFC rating. The three other misclassifications (cnet, cnn, and thetruereporter) show the same pattern.
\subsubsection{Examples of corrected errors after using temporal features with BD}
\label{bd_temp_err}
One of the domains misclassified before applying temporal weighting is medium.com, which includes 1 neutral article and 3 biased ones. This results in a biased proportion of 0.75 (3/4) and a neutral proportion of approximately 0.25, leading to the domain being classified as biased. However, when temporal weighting is introduced, the classification shifts. The neutral article—titled “Steve Scalise Says Ady Barkan asked Joe Biden, ‘Do we agree that we can redirect some of the funding for police?’”—was published in 2020 and thus receives a higher temporal weight (0.4835). In contrast, the three biased articles were published in 2016, resulting in a combined weight of 0.5616. This adjustment shifts the classification toward 'slightly\_biased'.
\subsubsection{Errors in credibility assessment of the domains using DCEF}
\label{kokobt}
As discussed in Experiment \ref{cred_exp}, two major patterns emerge among the 11 misclassifications (out of 91 test domains) in the credibility evaluation of web domains. The first pattern—hallucinations in the assessment of highly temporally weighted article features such as factuality, bias, propaganda, or title quality—was observed for domains including medium.com and thepostemail.com.

Notably, these errors result in only minor shifts in the predicted credibility levels, such as from low to medium or from medium to high, rather than more significant misclassifications (e.g., from low to high or vice versa).

The second pattern—errors arising from an insufficiently representative sample of articles from a given domain—is observed in the misclassifications of twitchy.com, foxnews.com, and facebook.com. In this case, two out of the three domains exhibit substantial shifts in their credibility assessments, being predicted as highly credible despite actually having low credibility.

This outcome is expected, as each domain is represented by only a single article, causing the evaluation to depend heavily on the features of that specific article rather than reflecting the domain’s overall characteristics.

\begin{figure}
  \centering
  \includegraphics[width=7.5cm]{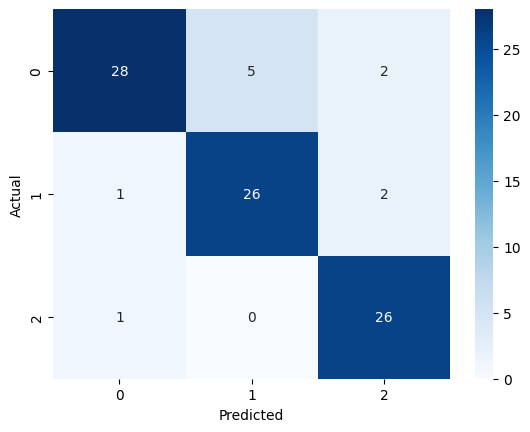}
  \caption{The confusion matrix on DECF predicted classes; low credibility (0), medium credibility (1), and high credibility (2) }
  \label{conf_mat_old}
\end{figure}

The remaining domains (guardianlv.com, unitynewsnetwork.co.uk, tamparepublic.com, upgazette.com, theregionnews.com, and act.tv) shows both patterns. Figure \ref{conf_mat_old} shows the confusion matrix for the 11 misclassifications.

\subsubsection{Errors in credibility assessment of the domains with new content}
\label{errrr}
Regarding the 60 domains containing newly published articles that were not seen by Llama3.3-70B, a total of five misclassifications were observed: metro.co.uk, sonsof1776.com, upgazette.com, thepostemail.com, and wcalabamanews.com. As illustrated in Figure \ref{conf_mat}, these misclassifications involve only moderate shifts in credibility (e.g., from low to medium or vice versa), rather than extreme errors.

Among these cases, the second error pattern—stemming from insufficiently representative article samples—is observed for metro.co.uk and thepostemail.com. The remaining misclassifications follow the first pattern, where errors arise due to hallucinations in the assessment of temporally high weighted article features.

The performance metrics in this experiment are higher than those in Experiment \ref{cred_exp} because the latter reports averages across five folds with varying performance, while this experiment uses a fixed data split.

\subsection{LLM Selection Experiments}
\label{LLM_sec_exp}

The goal of the following experiments is to evaluate multiple LLMs of varying sizes from different providers as potential underlying models for DCEF. LLMs were selected over other supervised methods and PLMs due to their superior ability to generalize across diverse topics and datasets \cite{srba2024survey}. In addition, unlike PLMs, LLMs can provide post-hoc justifications for their decisions.
\subsubsection{LLM selection for the Propaganda Detection Module (PD)}
Three different LLMs were evaluated on three datasets for the task of propaganda detection, with the goal of selecting the most effective model for the Propaganda Detector (PD) module. They are also compared against the SOTA DistilRoBERTa fine-tuned on Proppy \cite{barron2019proppy} in propaganda detection. The datasets include Proppy, PTC \cite{da2020semeval}, and the Memes Propaganda Detection dataset introduced by \citet{dimitrov2021detecting}. The three LLMs tested are: Llama-3.3-70B (Meta), Gemma3-27B (Google), and Phi-4-14B (Microsoft). The models except for DistilRoBERTa were CoT zero-shot prompted. The full prompt used for LLM-based models is provided in the Appendix \ref{app_prompts} Figure \ref{propaganda_propmt}.

As shown in Table \ref{propaganda_ev}, DistilRoBERTa achieves the highest F1-score on the Proppy dataset, which is expected given that it was trained on this dataset. However, its F1-score drops significantly on the Memes dataset and is only moderate on PTC. This suggests that while DistilRoBERTa represents the state of the art on Proppy, but its superior performance does not generalize well across datasets. In contrast, Llama-3.3-70B achieves the best F1-scores on both PTC and the Memes dataset, while ranking second on Proppy. This demonstrates its strong generalization capability across different datasets without fine-tuning. Interestingly, Gemma3-27B performs comparably to Llama-3.3-70B in terms of F1-scores across datasets, despite having significantly fewer parameters, highlighting its efficiency relative to larger models in propaganda detection. Llama-3.3-70B has been chosen for the PD module based on Table \ref{propaganda_ev} results as it has beaten the SOTA DistilRoBERTa \cite{barron2019proppy} in generalization and consistency.

\begin{table}
\centering
\resizebox{\columnwidth}{!}{%
\begin{tabular}{|c|c|c|c|c|}
\hline
 & distil-roberta & Llama-3.3-70B & Gemm3-27B & Phi-4-14B \\ \hline
Proppy & \textbf{0.92} & 0.82 & 0.81 & 0.72 \\ \hline
PTC & 0.71 & \textbf{0.72} & 0.70 & 0.61 \\ \hline
Memes & 0.62 & \textbf{0.84} & 0.82 & 0.59 \\ \hline
\end{tabular}%
}
\caption{Showing micro F1-scores for classification of propaganda of four different models against three different datasets.}
\label{propaganda_ev}
\end{table}

\subsubsection{LLM selection for the Bias Detector (BD).}
\citet{baly2020we} have created a balanced dataset with two classes (biased, neutral) for the purpose of political bias detection in articles. They tried fine-tuning BERT with triplet loss and adversarial adaptation for de-biasing the effect of the domain on the article-level bias detection. They achieved an accuracy of 51.41\% on their dataset. For the purpose of comparison, we use accuracy as the performance metric, and five LLMs were zero-shot prompted for bias detection and the prompt is in Figure \ref{bias_propmt}. As shown in Table \ref{bias_ev}, Llama-3.3-70B has the highest accuracy surpassing all the models in bias detection and that is why it is chosed as the underlying LLM for BD. 
 
\begin{table}
\centering
\resizebox{\columnwidth}{!}{%
\begin{tabular}{c|c|c|c|c|c|}
\cline{2-6}
 & \begin{tabular}[c]{@{}c@{}}Llama-3.3-\\ 70B\end{tabular} & \begin{tabular}[c]{@{}c@{}}Gemm3-\\ 27B\end{tabular} & \begin{tabular}[c]{@{}c@{}}Phi-4-\\ 14B\end{tabular} & \begin{tabular}[c]{@{}c@{}}Qwen-\\ 32B\end{tabular} & \begin{tabular}[c]{@{}c@{}}Deepseek-r1\\ -70B\end{tabular} \\ \hline
\multicolumn{1}{|c|}{Acc} & \textbf{0.64} & 0.58 & 0.51 & 0.46 & 0.60 \\ \hline
\end{tabular}%
}
\caption{Showing accuracies for classification of bias of five different models.}
\label{bias_ev}
\end{table}
This is consistent with the findings of \citet{sar2024navigating}. They proved that using CoT and in-context learning made Llama3-70B
achieve comparable results with the SOTA ConvBert on the MMIB benchmark \cite{wessel2023introducing}. 
Llama-3.3-70B has been chosen after beating the SOTA model by \citet{baly2020we}.

\subsubsection{LLM selection for the Factuality Evaluation Module (FEM)}
To evaluate the capability of the factuality evaluation module, we randomly sampled 508 records from the PolitiFact dataset \cite{garg2020new}. This dataset classifies claims into six labels. For the purpose of our study, the original six labels in the dataset were consolidated into three categories. Specifically, the labels mostly-false, false, and pants-fire were grouped into misleading; the labels true and mostly-true were grouped into factual; and the label half-true was renamed as mixed. The politifact sample contains 292 misleading claims, 101 factual and 92 mixed. We tried 5 different LLMs as shown in Table \ref{factt_ev}.
\begin{table}
\centering
\resizebox{\columnwidth}{!}{%
\begin{tabular}{c|c|c|c|c|c|}
\cline{2-6}
 & \begin{tabular}[c]{@{}c@{}}Llama-3.3-\\ 70B\end{tabular} & \begin{tabular}[c]{@{}c@{}}Gemm3-\\ 27B\end{tabular} & \begin{tabular}[c]{@{}c@{}}Phi-4-\\ 14B\end{tabular} & \begin{tabular}[c]{@{}c@{}}Qwen-\\ 32B\end{tabular} & \begin{tabular}[c]{@{}c@{}}Deepseek-r1\\ -70B\end{tabular} \\ \hline
\multicolumn{1}{|c|}{\begin{tabular}[c]{@{}c@{}}F1-\\ score\end{tabular}} & \textbf{0.91} & 0.88 & 0.81 & 0.77 & 0.89 \\ \hline
\end{tabular}%
}
\caption{Showing micro F1-score for classification of article veracity of five different models. Llama-3.3-70B showed the highest F1-score.}
\label{factt_ev}
\end{table}
The F1-score achieved by the Llama (FEM) is 91\%. This performance encouraged us to use the Llama FEM in our following experiments. 

\subsection{Making sure that the underlying LLM (Llama3.3) used in DCEF can not guess the source web domain from the article text}
\label{guess_dom}

From Section \ref{LLM_sec_exp}, the top-performing model is Llama3.3-70B; therefore, all subsequent experiments focus on this LLM, which was released in December 2024. In this experiment, we assess whether Llama3.3-70B can correctly identify the domain of 60 domains containing content published after December 2024. This ensures that DCEF evaluates only genuinely new content that was not seen during the model’s training.

This experiment is designed as a baseline for determining whether DCEF can reliably assess domains that contain previously unseen content.

Setup: We randomly selected one article from each of the 60 domains, comprising 30 newly emerging domains (introduced after the release of Llama3.3) and 30 established domains, where the selected articles were published after the model’s release date. We then prompted Llama3.3-70B with the following instruction: “You are given the following article from an unknown web domain. Based on your knowledge, guess which domain this article belongs to ( the source web domain of the article). Restrict your output to the domain name only without any explanation.” We repeated this experiment for three times and each time a random article is chosen per domain.

Results: LLama3.3 was unable to guess the correct domains in the three runs as half of these domains (30 domains) were likely unseen by Llama3.3 and the other half's content is totally new for the LLM.

It is observed that, when presented with unseen content, Llama3.3-70B tends to infer domains whose names are semantically similar to the article’s content. For example, given the article: “European Union leaders meeting at the 2026 summit agreed to deepen defense cooperation in response to ongoing security concerns along the bloc’s eastern borders and shifting global geopolitical conditions.” the model predicted the domain “euroactiv.com,” whereas the true source was “truththeory.com.”

Similarly, for the article: “Salmon Pasta Recipe. Combine salmon with creamy cheese and let it melt on your palate—add pasta and a beach and it’ll taste as good as a holiday.” the model predicted “vincenzosplate.com,” while the actual domain was “diply.com.”

These examples illustrate that for new previously unseen content the model relies on topical cues and lexical associations rather than prior exposure to the specific source, further supporting the assumption that the content is indeed unseen.
\subsection{Modules Prompts}
\label{app_prompts}

All models were quantized (Q4\_K\_M) to address GPU RAM limitations. Models management and prompting were conducted using OLlama\footnote{\url{https://oLlama.com/}}.
For every module that relies on LLMs as its backbone—such as FEM, DTD, PD, and BD—a dedicated prompt is defined. In this section, we present the prompts used across these modules.

Political bias detection refers to the task of identifying whether a piece of content—such as news articles, social media posts, or political speeches—shows a partisan leaning toward a particular ideology, party, or perspective. It typically involves analyzing linguistic cues, framing, and patterns in how issues are presented. Below is the prompt used for political bias detection in the BD module:
\begin{figure}
\fcolorbox{blue}{gray!15}{%
    \begin{minipage}{\columnwidth}
\scriptsize
You are a neutral and rigorous political discourse analyst. Your task is to read a given article and determine whether its overall tone or framing leans left‑wing, center, or right‑wing.\\\\

I want you to analyze a text to detect political bias using a step-by-step reasoning process. Follow these steps:\\

1. Identify key language or framing cues—such as loaded terms, ideological keywords, framing of issues, quoted sources, and detected omissions.

2. For each cue, note whether it is typically associated with left‑leaning, center, or right‑leaning discourse (or neutral).

3. Summarize how these cues combine: do they overwhelmingly lean in one direction, remain balanced, or conflict?

4. Assess the strength of evidence: is the framing overt, subtle, or ambiguous?

5. Provide your final classification: “Left”, “Center”, or “Right”, and a brief rationale grounded in steps 1–4, as a short paragraph.

6. Your output is a json object containing only two fields which are the label and the explanation. Here is an example for your output:
{'Label':'biased or neutral', 'Explanation': 'your explanation'}

7. Do not add the instructions or anything from the prompt. Your output is restricted to the json object containing the label and the explanation only.\\

Now, apply this process to the following text:

article:\\
\{article\}
 \end{minipage}%
}
  \caption{Prompt used for political bias detection.}
  \label{bias_propmt}

\end{figure}

Credible articles typically have titles that align with their content. When a title is deceptive or contradictory to the article’s body, it often signals low credibility. The following prompt is used for deceptive article detection:

\begin{figure}
\fcolorbox{blue}{gray!15}{%
    \begin{minipage}{\columnwidth}
\scriptsize
You will be given the title of an article and the full text of the article. Your task is to determine whether the title is deceptive clickbait or not clickbait, based on whether it misrepresents or exaggerates the actual content of the article.\\\\

Instructions:\\

1. Analyze whether the title accurately reflects the article content.

2. Consider if the title is misleading, sensationalized, or uses curiosity gaps (e.g., "You won't believe what happened next").

3. A deceptive title often over-promises or omits key context to attract clicks, without delivering in the article.

4. An honest title should honestly summarize or reflect the content.

5. Your output should be in the following json format:
\{"Label": "deceptive or honest", "Justification": "Your explanation here"\}\\\\

Here is an example:

Title: "Scientists Have Finally Found a Cure for Aging!"

Article: [Full article text about a promising but early-stage anti-aging drug trial in mice]

the output:
\{"Label": "Deceptive",
"Justification": "The title implies a definitive cure for aging has been found, which is misleading. The article discusses early research in animals, not a proven cure."\}\\

Now here is the input you should evaluate:\\

Title:\\
\{title\}\\

article:\\
\{article\}
 \end{minipage}%
}
  \caption{Prompt used for deceptive titles detection.}
  \label{deceptive_propmt}

\end{figure}

Fact-checking generally involves two main stages. First, trustworthy information (evidence) relevant to a claim or article is retrieved. Then, the claim or article is evaluated based on its alignment with the retrieved evidence. The following prompt is used for evidence retrieval, while the subsequent one is for alignment checking.
\begin{figure}
\fcolorbox{blue}{gray!15}{%
    \begin{minipage}{\columnwidth}
\scriptsize
Provide me with information about the following article:\\
\{article\}\\
 \end{minipage}%
}
  \caption{Prompt used for evidence retrieval from Google search bar.}
  \label{evidence_propmt}

\end{figure}

\begin{figure}
\fcolorbox{blue}{gray!15}{%
    \begin{minipage}{\columnwidth}
\scriptsize
You are an expert media credibility analyst. Your task is to evaluate the credibility of \{domain\_name\} based on the content of a provided articles. Provide me with a corresponding list of credibility ratings for:\\
\{article\_list\}\\
 \end{minipage}%
}
  \caption{Prompt used by GPT-5.5 to evaluate domain credibility.}
  \label{gpt_cred}

\end{figure}

\begin{figure}
\fcolorbox{blue}{gray!15}{%
    \begin{minipage}{\columnwidth}
\scriptsize
You are a senior fact-checker. You are provided with an article and with reference trustworthy information called the evidence.\\ 

Check the alignment between the evidence "the reference" and the article. 

-If the article is completely aligned with the evidence, then the article is factual.

-If the article is partially aligned with the evidence, the the article is mixed.

-If the article is not in alignment with the evidence, then the article is misleading.

Your output should be a label presenting your judgment on the article with justification. Your output should follow the following json form:

\{'Label':'factual, mixed or misleading', 'Explanation': 'your explanation'\}

Article:\\
\{article\}\\

Evidence:\\
\{evidence\}\\
 \end{minipage}%
}
  \caption{Prompt used for fact-checking the article.}
  \label{fact_check_propmt}

\end{figure}

For propaganda detection, the prompt incorporates 18 propaganda techniques as defined in \cite{da2020semeval}:
\begin{figure}
\fcolorbox{blue}{gray!15}{%
    \begin{minipage}{\columnwidth}
\scriptsize
You are a senior media-literacy analyst. Your job is to determine whether a given article uses propaganda techniques. Think carefully and systematically. Do your detailed reasoning silently; only output the requested structured fields.\\
follow the following steps and think silently:\\

1. Comprehend the article: main claims, targets, and calls-to-action.

2. Separate facts vs. opinions; check whether claims are supported.

3. Assess tone (neutral vs. emotional/manipulative).

4.Scan for propaganda techniques (see taxonomy below).

5.Evaluate intent (inform vs. persuade vs. 
manipulate) and likely impact.

6.Form a calibrated judgment.\\\\

Propaganda Technique Taxonomy (detect any that apply):\\
-For each technique, match with exact text spans and a brief explanation.

-Loaded Language – emotionally charged wording to sway feelings rather than reason.

-Name-Calling / Labeling – attaching pejorative or glamorous labels to people/groups/ideas.

-Repetition – repeating key phrases/slogans to imprint a message.

-Exaggeration / Minimization – overstating positives/negatives or downplaying opposing facts.

-Doubt – seeding uncertainty about opponents, institutions, or evidence without solid grounds.

-Appeal to Fear / Prejudice – arousing fear or bias to push a viewpoint or action.

-Flag-Waving – linking a stance to patriotism, national duty, or group loyalty.

-Causal Oversimplification – claiming a single/simple cause or cure for complex issues.

-Slogans – short, memorable catchphrases used as proof or to replace argument.

-Appeal to Authority – citing authority/expert/celebrity as proof, without adequate evidence or outside their expertise.

-Black-and-White Fallacy – forcing a false “either/or” choice; ignoring nuance/alternatives.

-Thought-Terminating Clichés – stock phrases that shut down inquiry (e.g., “It is what it is”).

-Whataboutism – deflecting criticism by pointing to other wrongs instead of addressing the claim.

-Reductio ad Hitlerum – smearing by comparing to Nazis/Hitler to discredit without argument.

-Red Herring – introducing distractions/irrelevancies to derail or avoid the issue.

-Bandwagon – asserting “everyone agrees/is doing this” to pressure conformity.

-Obfuscation / Intentional Vagueness / Confusion – unclear terms, passive voice, or muddled info that hides meaning or responsibility.

-Straw Man – distorting an opposing view to attack the weaker version.\\

7. Provide your final classification: “propaganda”, or “neutral”, and a brief rationale grounded in steps 1–6, as a short paragraph.

8. Your output is a json object containing only three fields which are the label, and the explanation. Here is an example for your output:
\{'Label':'propaganda, or neutral', 'Explanation': 'your explanation'\}

9. Do not add the instructions or anything from the prompt. Your output is restricted to the json object containing the label and the explanation only.\\

Now, apply this process to the following text:\\
the article:\\
\{article\}
 \end{minipage}%
}
  \caption{Prompt used for propaganda detection.}
  \label{propaganda_propmt}

\end{figure}

\subsection{The Credibility Classifier}
\label{D_tree}
\begin{figure*}
  \centering
  \includegraphics[width=\textwidth]{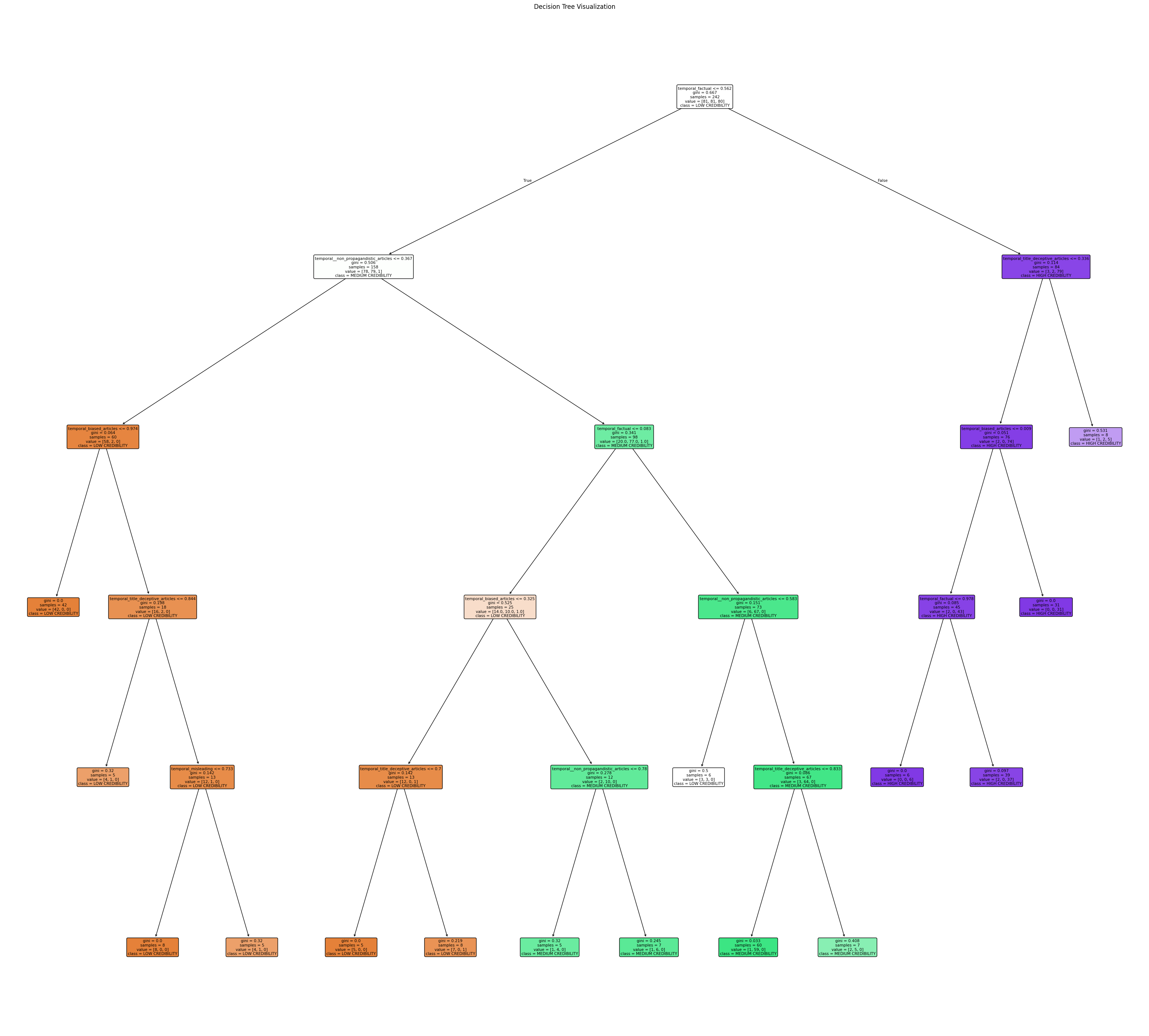}
  \caption{The final classifier for domain credibility classification when all the new 60 domains were placed in the test set.}
  \label{cred_tree}
\end{figure*}

In this section, we present the 5-level decision tree classifier that classified the 60 domains (domains that contain unseen content by Llama3.3) credibility in Figure \ref{cred_tree}.

\subsection{Illustrative Examples of LLMs Credibility Ratings for new Emerging Domains}
\label{cred_examples}
In this section, we present illustrative examples demonstrating how LLMs assess the credibility of newly emerged domains that they have not encountered before and for which they lack prior knowledge.
\subsubsection{LLAMA-3.3 rating examples}
\begin{enumerate}
  
 \item Regrading the \textbf{low} credibility domain gooddaytemple.com as per MBFC, here is the output of LLAMA:
 
 \textbf{LLAMA Label:} Low
 
 \textbf{Justification of LLAMA:} I rate the credibility of gooddaytemple.com as low because I couldn't find any information about this website being a reputable or well-established source. There is no readily available data on its track record, credentials, or expertise in any particular field. Without a clear understanding of the website's purpose, authorship, or peer review process, it's challenging to assess its credibility. Additionally, the domain name "gooddaytemple" seems vague and doesn't provide any immediate indication of the website's focus or authority. As a result, I would approach information from this website with skepticism and verify any claims through more trustworthy sources.
 
 \textbf{Our Analysis:} LLAMA rated this domain credibility as low based on the fact that it doesn't have enough information about it. This is not a reliable result despite the fact that the label is correct.

\item Regrading the \textbf{low} credibility domain migrationcenter.co.uk as per MBFC, here is the output of LLAMA:

\textbf{LLAMA Label}: Medium 

\textbf{Justification of LLAMA:} I rate the credibility of migrationcenter.co.uk as medium because it appears to be a website related to immigration and visa services, which is a niche area of expertise. While I couldn't find any prominent red flags, I also couldn't verify any notable credentials, certifications, or endorsements from reputable organizations. The website's domain name suggests a focus on UK immigration, which could indicate some level of specialization.

 \textbf{Our Analysis:} LLAMA rated this domain credibility based on an assumption (it appears to be a website related to immigration and visa services) that is not correct leading to an error in its label.

\item Regrading the \textbf{high} credibility domain witneygazette.com as per MBFC, here is the output of LLAMA:

 \textbf{LLAMA Label:} Medium
 
 \textbf{Justification of LLAMA:} I rate the credibility of witneygazette.com as medium because it appears to be a local news website serving the Witney area in Oxfordshire, UK. The website's domain name suggests a focus on local news, sports, and events, which is a relatively narrow scope.
 
 \textbf{Our Analysis:} LLAMA rated this domain credibility based on an assumption (that it is a local news website) that is not correct leading to an error in its label.

\item Regrading the \textbf{high} credibility domain bluegrassinstitute as per MBFC, here is the output:

\textbf{LLAMA Label:} Low

 \textbf{Justification of LLAMA:} I rate the credibility of bluegrassinstitute.org as low because it appears to be a think tank with a clear ideological bias, which can compromise its objectivity and credibility.
 
 \textbf{Our Analysis:} LLAMA is hallucinating.

\end{enumerate}
\subsubsection{Gemma3-27B Credibility Rating Examples}
\begin{enumerate}

\item Regrading the \textbf{low} credibility domain gooddaytemple.com as per MBFC, here is the output of Gemma:

\textbf{Gemma Label:} Medium

\textbf{Justification:} What the Site Is \& Its Purpose: gooddaytemple.com is the official website for Good Day Temple, a non-profit organization based in Temple City, California. They focus on providing resources and support (food, clothing, assistance programs) to low-income individuals and families in the San Gabriel Valley. This is a known, established charity.
Professionalism \& Transparency: The website has a fairly professional design, clear navigation, and contains a good amount of information about the organization’s mission, programs, and impact. They publicly list their address, contact information, and board members. Financial information (annual reports) is available on the site, which supports transparency.
External Validation (Positive): Good Day Temple is listed on GuideStar, a reputable database for non-profit organizations. They also have a reasonable rating (currently 4 stars) on Charity Navigator. These external validations increase credibility.
 \textbf{Our Analysis:} Gemma is hallucinating and providing false information in the justification.

\item Regrading the \textbf{low} credibility domain migrationcenter.co.uk as per MBFC, here is the output of Gemma:

\textbf{Gemma Label:} Medium

\textbf{Justification:} What the Site Is \& Its Purpose: migrationcenter.co.uk presents itself as a provider of information and support services related to UK immigration and visa applications. They offer guides, checklists, and seemingly some consultancy/application assistance services.
Professionalism: The website has a reasonably professional appearance, with clear navigation and a decent level of detail in some of the information provided. However, the design isn’t particularly modern and feels somewhat generic.
Lack of Clear Credentials \& Regulation: This is the biggest factor holding the credibility back. While they state they offer "expert advice," the site lacks clear, readily visible information about the qualifications and regulatory status of their advisors. Immigration advice in the UK is heavily regulated; advisors must be registered with the Office of the Immigration Services Commissioner (OISC). 

\textbf{Our Analysis:} Gemma is hallucinating and providing false information in the justification.

\item Regrading the \textbf{low} credibility domain quantisophy as per MBFC, here is the output of Gemma:

\textbf{Gemma Label:} High

\textbf{Justification:} What the Site Is \& Its Purpose: quantisophy.com is the website of Dr. Laurent Balmès, a quantitative analyst (quant) specializing in volatility trading and risk management. The site offers educational content (articles, videos) primarily focused on financial markets, options trading, and statistical analysis. It also offers a premium subscription service with more in-depth content.
Expertise \& Credentials: Dr. Balmès has a strong academic background (Ph.D. in theoretical physics) and extensive experience working as a quantitative analyst in the financial industry (specifically, volatility trading at Societe Generale). This background lends significant weight to his analysis. He openly details his qualifications and experience on the site.
Focus on Methodology \& Transparency: A key strength of Quantisophy is its emphasis on the underlying quantitative methodologies. Dr. Balmès clearly explains his models and the statistical principles behind his analyses. While the math can be complex, he attempts to demystify it.

\textbf{Our Analysis:} Gemma is hallucinating and providing false information in the justification.
\end{enumerate}

\subsection{Interpretability and Explainability}

A distinction should be made between interpretability and explainability, as these terms refer to related but conceptually different properties of a system. Interpretability refers to how well a human can understand the internal mechanics of a model. It describes how a model arrives at a decision based on its architecture and logic. Explainability refers to the ability to provide a post-hoc (after-the-fact) reason or justification for a specific output. 

For this reason, we argue that although the system may not be fully interpretable at the parameter level, it provides insights through its justifications at the architectural level. All the pipeline intermediate modules provide justifications for their classifications and the last classifier is a decision tree which is interpretable. It is important to note that we are not evaluating the quality of the justifications produced by each model. Our objective is not to generate high-quality justifications, but rather to produce justifications that provide insight into each decision made by DCEF. Improving the quality of the justifications is left for future work.

\subsection{Credibility and Factuality}

Credibility pertains to the trustworthiness or authority of a source. A person can be credible because they have a PhD or a long history of honesty, which makes you perceive their message as true. On the other hand, factuality is the degree to which information aligns with objective reality. It is possible for a credible person (an honest expert) to be wrong in a specific instance (a clouded memory), leading to a lack of factuality in their statement. A highly credible source must be frequently factual.
\subsection{Human Evaluations on article misinformation, bias, propaganda, and deceptive articles detection}
\label{hum_ann}
To assess the article-level evaluations generated by the LLMs, we will compare their assessments with human evaluations and measure the level of agreement between them.

Three human annotators (postgraduate students in computer science) were provided with 100 articles (web pages) along with reference information (evidence) to support factuality assessment. They were asked to determine whether each article contained bias, propaganda, deceptive titles, and/or misinformation. Their majority vote is then compared against the LLAMA-DCEF article evaluations on bias, propaganda, misinformation and deceptive titles detection.

The correlation between annotators is reported using Pearson correlation in Table \ref{title} for deceptive title detection labels, Table \ref{bias} for bias detection, Table \ref{propag} for propaganda detection, and Table \ref{factuality_tab} for misinformation detection.

\begin{table}
\centering
\resizebox{\columnwidth}{!}{%
\begin{tabular}{
>{\columncolor[HTML]{FFFFFF}}c |
>{\columncolor[HTML]{FFFFFF}}c |
>{\columncolor[HTML]{FFFFFF}}c |
>{\columncolor[HTML]{FFFFFF}}c |}
\cline{2-4}
{\color[HTML]{1F1F1F} \textbf{}} & {\color[HTML]{1F1F1F} \textbf{ann\_1\_title\_label}} & {\color[HTML]{1F1F1F} \textbf{ann\_2\_title\_label}} & {\color[HTML]{1F1F1F} \textbf{ann\_3\_title\_label}} \\ \hline
\multicolumn{1}{|c|}{\cellcolor[HTML]{FFFFFF}{\color[HTML]{1F1F1F} \textbf{ann\_1\_title\_label}}} & {\color[HTML]{1F1F1F} 1.000000} & {\color[HTML]{1F1F1F} 0.921597} & {\color[HTML]{1F1F1F} 0.899376} \\ \hline
\multicolumn{1}{|c|}{\cellcolor[HTML]{FFFFFF}{\color[HTML]{1F1F1F} \textbf{ann\_2\_title\_label}}} & {\color[HTML]{1F1F1F} 0.921597} & {\color[HTML]{1F1F1F} 1.000000} & {\color[HTML]{1F1F1F} 0.903200} \\ \hline
\multicolumn{1}{|c|}{\cellcolor[HTML]{FFFFFF}{\color[HTML]{1F1F1F} \textbf{ann\_3\_title\_label}}} & {\color[HTML]{1F1F1F} 0.899376} & {\color[HTML]{1F1F1F} 0.903200} & {\color[HTML]{1F1F1F} 1.000000} \\ \hline
\end{tabular}%
}
\caption{Pearson correlation between the three human annotators on labeling whether the article title is decpetive or not}
\label{title}
\end{table}

\begin{table}
\centering
\resizebox{\columnwidth}{!}{%
\begin{tabular}{
>{\columncolor[HTML]{FFFFFF}}c |
>{\columncolor[HTML]{FFFFFF}}c |
>{\columncolor[HTML]{FFFFFF}}c |
>{\columncolor[HTML]{FFFFFF}}c |}
\cline{2-4}
\multicolumn{1}{r|}{\cellcolor[HTML]{FFFFFF}{\color[HTML]{1F1F1F} \textbf{}}} & {\color[HTML]{1F1F1F} \textbf{ann\_1\_bias\_label}} & {\color[HTML]{1F1F1F} \textbf{ann\_2\_bias\_label}} & {\color[HTML]{1F1F1F} \textbf{ann\_3\_bias\_label}} \\ \hline
\multicolumn{1}{|c|}{\cellcolor[HTML]{FFFFFF}{\color[HTML]{1F1F1F} \textbf{ann\_1\_bias\_label}}} & {\color[HTML]{1F1F1F} 1.000000} & {\color[HTML]{1F1F1F} 0.815951} & {\color[HTML]{1F1F1F} 0.657129} \\ \hline
\multicolumn{1}{|c|}{\cellcolor[HTML]{FFFFFF}{\color[HTML]{1F1F1F} \textbf{ann\_2\_bias\_label}}} & {\color[HTML]{1F1F1F} 0.815951} & {\color[HTML]{1F1F1F} 1.000000} & {\color[HTML]{1F1F1F} 0.805354} \\ \hline
\multicolumn{1}{|c|}{\cellcolor[HTML]{FFFFFF}{\color[HTML]{1F1F1F} \textbf{ann\_3\_bias\_label}}} & {\color[HTML]{1F1F1F} 0.657129} & {\color[HTML]{1F1F1F} 0.805354} & {\color[HTML]{1F1F1F} 1.000000} \\ \hline
\end{tabular}%
}
\caption{Pearson correlation between the three human annotators on labeling whether the article is biased or not}
\label{bias}
\end{table}

\begin{table}
\centering
\resizebox{\columnwidth}{!}{%
\begin{tabular}{
>{\columncolor[HTML]{FFFFFF}}c |
>{\columncolor[HTML]{FFFFFF}}c |
>{\columncolor[HTML]{FFFFFF}}c |
>{\columncolor[HTML]{FFFFFF}}c |}
\cline{2-4}
\multicolumn{1}{r|}{\cellcolor[HTML]{FFFFFF}{\color[HTML]{1F1F1F} \textbf{}}} & {\color[HTML]{1F1F1F} \textbf{ann\_1\_propaganda}} & {\color[HTML]{1F1F1F} \textbf{ann\_2\_propaganda}} & {\color[HTML]{1F1F1F} \textbf{ann\_3\_propaganda}} \\ \hline
\multicolumn{1}{|c|}{\cellcolor[HTML]{FFFFFF}{\color[HTML]{1F1F1F} \textbf{ann\_1\_propaganda}}} & {\color[HTML]{1F1F1F} 1.000000} & {\color[HTML]{1F1F1F} 0.854135} & {\color[HTML]{1F1F1F} 0.652141} \\ \hline
\multicolumn{1}{|c|}{\cellcolor[HTML]{FFFFFF}{\color[HTML]{1F1F1F} \textbf{ann\_2\_propaganda}}} & {\color[HTML]{1F1F1F} 0.854135} & {\color[HTML]{1F1F1F} 1.000000} & {\color[HTML]{1F1F1F} 0.763510} \\ \hline
\multicolumn{1}{|c|}{\cellcolor[HTML]{FFFFFF}{\color[HTML]{1F1F1F} \textbf{ann\_3\_propaganda}}} & {\color[HTML]{1F1F1F} 0.652141} & {\color[HTML]{1F1F1F} 0.763510} & {\color[HTML]{1F1F1F} 1.000000} \\ \hline
\end{tabular}%
}
\caption{Pearson correlation between the three human annotators on labeling whether the article contains propaganda or not}
\label{propag}
\end{table}

\begin{table}
\centering
\resizebox{\columnwidth}{!}{%
\begin{tabular}{
>{\columncolor[HTML]{FFFFFF}}c |
>{\columncolor[HTML]{FFFFFF}}c |
>{\columncolor[HTML]{FFFFFF}}c |
>{\columncolor[HTML]{FFFFFF}}c |}
\cline{2-4}
{\color[HTML]{1F1F1F} \textbf{}} & {\color[HTML]{1F1F1F} \textbf{ann\_1\_fact\_check}} & {\color[HTML]{1F1F1F} \textbf{ann\_2\_fact\_check}} & {\color[HTML]{1F1F1F} \textbf{ann\_3\_fact\_check}} \\ \hline
\multicolumn{1}{|c|}{\cellcolor[HTML]{FFFFFF}{\color[HTML]{1F1F1F} \textbf{ann\_1\_fact\_check}}} & {\color[HTML]{1F1F1F} 1.000000} & {\color[HTML]{1F1F1F} 0.887997} & {\color[HTML]{1F1F1F} 0.812819} \\ \hline
\multicolumn{1}{|c|}{\cellcolor[HTML]{FFFFFF}{\color[HTML]{1F1F1F} \textbf{ann\_2\_fact\_check}}} & {\color[HTML]{1F1F1F} 0.887997} & {\color[HTML]{1F1F1F} 1.000000} & {\color[HTML]{1F1F1F} 0.914920} \\ \hline
\multicolumn{1}{|c|}{\cellcolor[HTML]{FFFFFF}{\color[HTML]{1F1F1F} \textbf{ann\_3\_fact\_check}}} & {\color[HTML]{1F1F1F} 0.812819} & {\color[HTML]{1F1F1F} 0.914920} & {\color[HTML]{1F1F1F} 1.000000} \\ \hline
\end{tabular}%
}
\caption{Pearson correlation between the three human annotators on labeling whether the article are factual, mixed or misleading.}
\label{factuality_tab}
\end{table}

\begin{table}
\centering
\resizebox{\columnwidth}{!}{%
\begin{tabular}{c|c|c|c|c|}
\cline{2-5}
\cellcolor[HTML]{FFFFFF}{\color[HTML]{1F1F1F} \textbf{}} & \cellcolor[HTML]{FFFFFF}{\color[HTML]{1F1F1F} \textbf{\begin{tabular}[c]{@{}c@{}}Bias\\ Detection\end{tabular}}} & \cellcolor[HTML]{FFFFFF}{\color[HTML]{1F1F1F} \textbf{\begin{tabular}[c]{@{}c@{}}Deceptive title\\ Detection\end{tabular}}} & \textbf{\begin{tabular}[c]{@{}c@{}}Propaganda\\ Detection\end{tabular}} & \cellcolor[HTML]{FFFFFF}{\color[HTML]{1F1F1F} \textbf{\begin{tabular}[c]{@{}c@{}}misinformation\\ detection\end{tabular}}} \\ \hline
\multicolumn{1}{|c|}{\cellcolor[HTML]{FFFFFF}{\color[HTML]{1F1F1F} \textbf{Correlation}}} & \cellcolor[HTML]{FFFFFF}{\color[HTML]{1F1F1F} 0.816} & \cellcolor[HTML]{FFFFFF}{\color[HTML]{1F1F1F} 0.960} & 0.854 & \cellcolor[HTML]{FFFFFF}{\color[HTML]{1F1F1F} 0.888} \\ \hline
\multicolumn{1}{|c|}{\textbf{p-value}} & 4.670e-25 & 4.243e-56 & 1.395e-29 & 7.798e-35 \\ \hline
\end{tabular}%
}
\caption{Pearson correlation between the DCEF framework using llama3.3-70B as its underlying LLM and the majority vote of the three human annotatos in four different tasks.}
\label{final_corr}
\end{table}

It should be noted that the average inter-annotator correlation is 0.90 for deceptive title detection, 0.76 for bias detection, 0.75 for propaganda detection, and 0.87 for misinformation detection. 

Deceptive title detection achieves the highest correlation because it is a relatively straightforward task that primarily relies on assessing the alignment between the article title and the article content. Misinformation detection shows the second-highest correlation, as it depends on evaluating the consistency between the article and the reference evidence. 

In contrast, bias detection is more complex because it does not involve a direct reference text for comparison. Finally, propaganda detection is the most challenging task among them, as it requires identifying 16 different propaganda techniques in order to determine the final label. These techniques are outlined in the instructions provided to the human annotators.

After obtaining the majority-vote labels for the four tasks from the three human annotators and comparing the resulting classifications, all Pearson correlation scores across the four tasks were above 0.8, as shown in Table \ref{final_corr}, which is considered significant in the social sciences. \citet{10.1093/geroni/igz036} note that while 0.8+ correlation is "strong" in a general sense, 0.5 and above is the threshold for a "strong" consistent pattern in typical behavioral research.

\paragraph{The annotation guidelines provided to the annotators are as follows:\\\\}

\textit{You are provided with an article, its title, and some trustworthy reference information called the evidence.
You have four different tasks:}

\paragraph{\textit{Political bias detection:}}
\textit{Analyze a text to detect political bias using a step-by-step process.}
\textit{\begin{itemize}
  \item Identify key language or framing cues—such as loaded terms, ideological keywords, framing of issues, quoted sources, and detected omissions.
  \item For each cue, note whether it is typically associated with left- leaning, center, or right-leaning discourse (or neutral).
  \item Summarize how these cues combine: do they overwhelmingly lean in one direction, remain balanced, or conflict?
  \item Provide your final classification: neutral or biased
\end{itemize}}

\paragraph{\textit{Deceptive titles detection:}}
\textit{Your task is to determine whether the title is deceptive clickbait or not clickbait, based on whether it misrepresents or exaggerates the actual content of the article:}
\textit{\begin{itemize}
  \item Analyze whether the title accurately reflects the article content.
  \item Consider if the title is misleading, sensationalized, or uses curiosity gaps (e.g., "You won’t believe what happened next").
  \item A deceptive title often over-promises or omits key context to attract clicks, without delivering in the article.
  \item An honest title should honestly summarize or reflect the content.
  \item Your output should be deceptive or honest
\end{itemize}}

\paragraph{Misinformation detection:}
\textit{Check the alignment between the evidence "the reference" and the article.}
\textit{   \begin{itemize}
     \item If the article is completely aligned with the evidence, then the article is factual.
     \item If the article is partially aligned with the evidence, the article is mixed.
     \item If the article is not in alignment with the evidence, then the article is misleading.
   \end{itemize}}
\textit{Your output should be a label presenting your judgment on the article: factual, mixed, or misleading.}

\paragraph{\textit{Propaganda detection:}}
\textit{Determine if any of the following techniques are used in the article or not:}

\textit{\begin{enumerate}
  \item Loaded Language – emotionally charged wording to sway feelings rather than Reason. Ex: Only a heartless bureaucrat would support this policy.” It uses emotional wording (“heartless”) to influence feelings instead of reasoning.
  \item Name-Calling / Labeling – attaching pejorative or glamorous labels to people/groups/ideas. Ex: “Don’t listen to her—she’s just another radical extremist.” It Attaches a negative label to discredit without argument.
  \item Repetition – repeating key phrases/slogans to imprint a message. Ex: “Our plan means jobs, jobs, jobs for everyone.” It repeats a phrase to make it stick in people’s minds.
  Exaggeration / Minimization – overstating positives/negatives or downplaying opposing facts. Ex: “It’s just a minor issue, nothing to worry about.” It overstates harm or downplaying significance.
  \item Doubt – seeding uncertainty about opponents, institutions, or evidence without solid grounds. Ex: “Can we really trust these scientists? What if they’re hiding something?”. It Seeds suspicion without providing real evidence.
  \item Appeal to Fear / Prejudice – arousing fear or bias to push a viewpoint or action. Ex: “If we don’t act now, our neighborhoods will become unsafe and overrun.”. It Triggers fear or bias to push action.
  Flag-Waving – linking a stance to patriotism, national duty, or group loyalty. Ex: “Support this policy—it’s the patriotic thing to do for our country.” It Links agreement with loyalty or national pride.  \item Causal Oversimplification – claiming a single/simple cause or cure for complex Issues. Ex: “Crime is rising because of one policy. Remove it, and everything will be fixed.”. It Reduces a complex issue to a single cause/solution.
  \item Slogans – short, memorable catchphrases used as proof Ex: “Take Back Control!” Short, catchy phrase presented as if it proves something.
  \item Appeal to Authority – citing authority/expert/celebrity as proof, without adequate evidence or outside their expertise. Ex: “A famous actor says this diet works, so it must be true.”. It Relies on authority outside their expertise instead of evidence.
  \item Black-and-White Fallacy – forcing a false “either/or” choice; ignoring nuance/alternatives. Ex: “You’re either with us, or you’re against us.”. It ignores any middle ground or nuance.
  \item Thought-Terminating Clichés – stock phrases that shut down inquiry (e.g., “It is what it is”). Ex: “That’s just the way things are.”. It Ends discussion instead of engaging with the issue.
  \item Whataboutism – deflecting criticism by pointing to other wrongs instead of addressing the claim. Ex: “You’re criticizing this policy, but what about what your side did last year?” It Deflects instead of addressing the criticism.
  \item Reductio ad Hitlerum – smearing by comparing to Nazis/Hitler to discredit without argument. Ex: “That policy is exactly what Hitler would have supported.” It Discredits by extreme comparison rather than reasoning.
  \item Red Herring – introducing distractions/irrelevancies to derail or avoid the issue. Ex: “We shouldn’t focus on pollution—what about the jobs this factory creates?” It Diverts attention to an unrelated issue.
  \item Bandwagon – asserting “everyone agrees/is doing this” to pressure conformity. Ex: “Everyone is switching to this product—why aren’t you?” It Pressures conformity by claiming widespread support.
  \item Obfuscation / Intentional Vagueness / Confusion – unclear terms, passive voice, or muddled info that hides meaning or responsibility. Ex: “Mistakes were made during the process.” It Avoids clarity and responsibility (who made the mistakes?).
  \item Straw Man – distorting an opposing view to attack the weaker version. Ex: Person A: “We should regulate social media to reduce harm.” Person B: “So you want to censor all free speech?” It Misrepresents the original argument to make it easier to attack.
\end{enumerate}
\textbf{Provide your final classification: “propaganda” or “neutral” }}

\end{document}